%% file: gdp-paper.tex
\documentclass[11pt]{article}
\pdfoutput=1
\usepackage{acl2010}
\usepackage{times}
\usepackage{latexsym}
\usepackage{amsmath}
\usepackage{url}

\usepackage{algorithm2e}
\usepackage{amssymb}
\usepackage{verbatim}
\usepackage{array}
\usepackage{tikz}
\usepackage{tikz-dependency}

\DeclareMathOperator*{\argmax}{arg\,max}
\newcommand\Mark[1]{\textsuperscript#1}

\title{A Bayesian Model for Generative Transition-based Dependency Parsing}
\author{Jan Buys\Mark{1} \and Phil Blunsom\Mark{1}\Mark{,}\Mark{2} \\
  \Mark{1}Department of Computer Science, University of Oxford 
  \Mark{2}Google DeepMind \\
  {\tt \{jan.buys,phil.blunsom\}@cs.ox.ac.uk} \\}
\date{}

\begin{document}
\maketitle

%max 200 words
\begin{abstract}
We propose a simple, scalable, fully generative model for transition-based 
dependency parsing with high accuracy.
The model, parameterized by Hierarchical Pitman-Yor Processes, overcomes the 
limitations of previous generative models by allowing fast and accurate inference. 
We propose an efficient decoding algorithm based on particle filtering
that can adapt the beam size to the uncertainty in the model while jointly
predicting POS tags and parse trees. 
The UAS of the parser is on par with that of a greedy discriminative baseline.
As a language model, it obtains better perplexity than a $n$-gram model
by performing semi-supervised learning over a large unlabelled corpus.
We show that the model is able to generate locally and syntactically coherent
sentences, opening the door to further applications in language generation. 
\end{abstract}

\section{Introduction}
\label{sec-introduction}

Transition-based dependency parsing algorithms 
that perform greedy local inference have proven to be very successful 
at fast and accurate discriminative parsing~\cite{Nivre08,ZhangN11,ChenM14}. 
Beam-search decoding further improves performance~\cite{ZhangC08,HuangS10,ChoiM13}, 
but increases decoding time. Graph-based 
parsers~\cite{McDonaldCP05,KooC10,LeiBJ14} perform global inference
and although they are more accurate in some cases, inference tends to be 
slower.  

In this paper we aim to transfer the advantages of transition-based
parsing to generative dependency parsing. 
While generative models have been used widely and
successfully for constituency parsing~\cite{Collins97,PetrovBTK06}, their
use in dependency parsing has been limited.
Generative models offer a principled approach to semi- and unsupervised
learning, and can also be applied to natural language
generation tasks. 

Dependency grammar induction models~\cite{KleinM04,BlunsomC10}
are generative, but not expressive enough for high-accuracy
parsing. % in either parsing or language modelling. 
A previous generative transition-based dependency parser~\cite{TitovH07} 
obtains competitive accuracies, but
training and decoding is computationally very expensive. 
Syntactic language models have also been shown to improve performance 
in speech recognition and machine translation~\cite{ChelbaJ00,CharniakKY03}. 
However, the main limitation of most existing generative syntactic models is
their inefficiency.

We propose a generative model for transition-based parsing (\S \ref{sec-transition-parsing}).
The model, parameterized by 
Hierarchical Pitman-Yor Processes (HPYPs)~\cite{Teh06}, 
learns a distribution over derivations of 
parser transitions, words and POS tags (\S \ref{sec-probability-models}). 

To enable efficient inference, we propose a novel algorithm for linear-time 
decoding in a generative transition-based parser (\S \ref{sec-inference}).
The algorithm is based on particle filtering~\cite{DoucetDG01}, a method 
for sequential Monte Carlo sampling. This method enables the beam-size during
decoding to depend on the uncertainty of the model.
%A particle filter is also used to sample derivations for semi- or unsupervised learning.

Experimental results (\S \ref{sec-experiments}) show that the model obtains $88.5\%$ UAS 
on the standard WSJ parsing task, compared to $88.9\%$ for a greedy discriminative model
with similar features. 
The model can accurately parse up to $200$ sentences per second.  
Although this performance is below state-of-the-art discriminative models, 
it exceeds existing generative dependency parsing models in either accuracy, speed
or both. 
%Analysis shows that the main limitation
%of the model lies in its limited ability to include lexical elements in the
%HPYP conditioning contexts: 
%When all lexical items are removed from the conditioning structure, performance
%only drops by $1\%$ UAS.

As a language model, the transition-based parser offers an inexpensive way
to incorporate syntactic structure into incremental word prediction. 
With supervised training the model's perplexity is comparable to that of $n$-gram models,
although generated examples shows greater syntactic coherence.
With semi-supervised learning over a large unannotated corpus its
perplexity is considerably better than that of a $n$-gram model.

\input{transition-parsing}

\input{probability-models}

\input{inference}

\input{experiments}

\input{related-work}

\section{Conclusion}
\label{sec-conclusion}

We presented a generative dependency parsing model that, unlike previous
models, retains most of the speed and accuracy of discriminative parsers. 
Our models can accurately estimate probabilities conditioned on
long context sequences. The model is scalable to large training and test 
sets, and even though it defines a full probability distribution over
sentences and parses, decoding speed is efficient.
Additionally, the generative model gives strong performance as a language
model. 
For future work we believe that this model can be applied successfully
to natural language generation tasks such as machine translation. 

%\section*{Acknowledgments}

\bibliographystyle{acl}
\bibliography{references}

\end{document}

%% file: transition-parsing.tex
\section{Generative Transition-based Parsing} %Non-projective
\label{sec-transition-parsing}

\begin{figure}
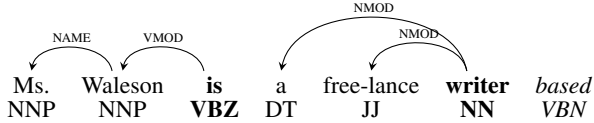

\small
\begin{dependency}[theme = simple]
\begin{deptext}[column sep=0.2cm]
    Ms. \& Waleson \& \textbf{is} \& a \& free-lance \& \textbf{writer} \& \emph{based} \\
    NNP \& NNP \& \textbf{VBZ} \& DT \& JJ \& \textbf{NN} \& \emph{VBN} \\
\end{deptext}
\depedge{2}{1}{NAME}
\depedge{3}{2}{VMOD}
\depedge{6}{5}{NMOD}
\depedge{6}{4}{NMOD}
\end{dependency} 
\caption{A partially-derived dependency tree for the sentence \emph{Ms. Waleson is a free-lance writer based in New York.} The next word to be predicted by the generative model is \emph{based}. 
     Words in bold are on the stack.}
\label{fig:deptree-conll}
\end{figure}

%\subsection{Transition-based dependency parsing}

Our parsing model is based on transition-based projective dependency parsing 
with the arc-standard parsing strategy~\cite{NivreS04}. 
%and arc-eager~\cite{Nivre03}
Parsing is restricted to (labelled) projective trees. 
An arc $(i, l, j) \in A$ encodes a dependency between two words,
where $i$ is the head node, $j$ the dependent and $l$ is the dependency type of 
$j$. 
In our generative model a word
can be represented by its lexical (word) type and/or its POS tag.
We add a root node to the beginning of the sentence (although it could
also be added at the end of the sentence), such that the head word of the sentence is the
dependent of the root node.

%%not using non-projective at the moment...
\begin{comment}
The non-projective extension follows the model of
\newcite{CohenGS11}, which is a variant of the system
proposed by \newcite{Attardi06}, permitting parsing of 
non-projective dependencies of transition degree $2$. 
This strategy has high coverage for non-projective dependencies that
occur in practice, without increasing the number of transition
steps needed to parse a sentence. 
\end{comment}

A parser configuration $(\mathbf{\sigma}, \beta, A)$ for sentence
$\mathbf{s}$ consists of a stack $\sigma$ of indices in $\mathbf{s}$, 
an index $\beta$ to the next word to be generated, and a set 
of arcs $A$. 
The stack elements are referred to 
as $\sigma_1, \ldots, \sigma_{|\sigma|}$, where $\sigma_1$ is the top element.
For any node $a$, $lc_1(a)$ refers to the leftmost child of $a$ in $A$, and 
$rc_1(a)$ to its rightmost child. 
%The second left-most and right-most children
%are referred to as $lc_2(a)$ and $rc_2(a)$, respectively. 

The initial configuration is $([], 0, \emptyset)$.
A terminal configuration is reached when $\beta > |\mathbf{s}|$,
and $\sigma$ consists only of the root. 
A sentence is generated left-to-right by performing
a sequence of transitions. % (its derivation). 
As a generative model
it assigns probabilities to sentences and dependency trees:
A word $w$ (including its POS tag) is generated when it is shifted on to the stack, 
similar to the generative models proposed by \newcite{TitovH07} and 
\newcite{CohenGS11}, and the joint tagging and parsing model of \newcite{BohnetN12}.

%\subsection{Arc-standard model}

%The first transition model used is the arc-standard transition system. 
%We make use of the arc-standard transition system. 
The types of transitions in this model are shift (sh), 
left-arc (la) and right-arc (ra):
\begin{itemize}
    \item[($\textrm{sh}_w$)] $(\sigma, i, A) \vdash (\sigma | i, i+1, A)$
    \item[($\textrm{la}_l$)] $(\sigma | i | j, \beta, A) \vdash (\sigma | j,
    \beta, A \cup \{ (j, l, i) \})$
    \item[($\textrm{ra}_l$)] $(\sigma | i | j, \beta, A) \vdash (\sigma | i,
    \beta, A \cup \{ (i, l, j) \})$
\end{itemize}

\begin{comment}
The non-projective transition system extends the projective system
by adding degree $2$ left-arc (la2) and right-arc (ra2) transitions:
\begin{itemize}
    \item[($\textrm{la2}_l$)] $(\sigma | i | j | k, \beta, A) \vdash 
                          (\sigma | j | k, \beta, A \cup \{ (k, l, i) \})$
    \item[($\textrm{ra2}_l$)] $(\sigma | i | j | k, \beta, A) \vdash 
                          (\sigma | i | j, \beta, A \cup \{ (i, l, k) \})$
\end{itemize}
\end{comment}

%say more about formal proporties of the system?

Left-arc and right-arc (reduce) transitions add an arc between the top two
words on the stack, and also generate an arc label $l$.
The parsing strategy adds arcs bottom-up.
No arc that would make the root node the dependent of another node may be added.
%Left-arc and right-arc transitions are referred to as reduce
%transitions, as they remove a node from the stack. 
%Each shift transition is associated with the generation of the word that is 
%shifted onto the stack, and each reduce transition with the label of the arc 
%that is added. 
To illustrate the generative process, the configuration of a partially generated
dependency tree is given in Figure \ref{fig:deptree-conll}.

In general parses may have multiple 
derivations. In transition-based parsing it is common to define an oracle $o(c, G)$ 
that maps the current configuration $c$ and the gold parse $G$ to the next 
transition that should be performed. In our probabilistic model we are interested
in performing inference over all latent structure, including spurious derivations. 
Therefore we 
propose a non-deterministic oracle which allows us to find all derivations of 
$G$. In contrast to dynamic oracles~\cite{GoldbergN13}, we are 
only interested in derivations of the correct parse tree, so the oracle can assume
that given $c$ there exists a derivation for $G$. 

First, to enforce the bottom-up property our oracle has to ensure that an arc
$(i, j)$ in $G$ may only be added once $j$ has been attached to all its children
-- we refer to these arcs as \emph{valid}. 
Most deterministic oracles add valid arcs greedily. 
Second, we note that if there exists a valid arc between $\sigma_2$ and $\sigma_1$ 
and the oracle decides to shift, the same pair will only occur on the top of the stack again after a
right dependent has been attached to $\sigma_1$. 
Therefore right arcs
have to be added greedily if they are valid, while adding a valid left arc may be
delayed if $\sigma_1$ has unattached right dependents in $G$. 
%extension for non-projective parsing

\begin{comment}
\subsection{Arc-eager model}

We also define the arc-eager transition system. The transition types in
this model also includes reduce (re). The transition are defined as follows:
\begin{itemize}
    \item[($\textrm{sh}_w$)] $(\sigma, i, A) \vdash (\sigma | i, i+1, A)$
    \item[($\textrm{la}_l$)] $(\sigma | i, j, A) \vdash (\sigma, j, A \cup \{ (j, l, i) \} )$
    \item[($\textrm{ra}_l$)] $(\sigma | i, j, A) \vdash (\sigma | i, j, A \cup \{ (i, l, j) \} )$
    \item[(re)] $(\sigma | i, \beta, A) \vdash (\sigma, \beta, A)$
\end{itemize}
%%change r0ght0arc
%update buffer

We require that any right-arc should be followed immediately by shift
(usually the shift is included in the definition of right-arc, but for our 
generative model we decompose it into two consecutive actions).
Additional preconditions are that for left-arc $i$ should not already have a 
head in $A$, while for reduce $i$ is required to already have a head. 
Left arcs are added bottom-up and right arcs are added top-down.
If an arc exists between $\sigma_1$ and $\beta$ it must be added eagerly. 
Note that as a generative model, arcs between $\beta$ and words to the left
of $\beta$ are generated before $\beta$. In contrast, for arc-standard a word
is always generated before any arcs are attached to it. 

Derivation al ambiguity may occur where both shift and reduce may lead to
a valid parse, in cases where the node at the top of the stack already has a
head and all its dependents in the gold parse. Our oracle prefers to reduce
is such cases. 
\end{comment}

%% file: probability-models.tex
\section{Probabilistic Generative Model}
\label{sec-probability-models}

%Next we define the two classes of probability models used,
%hierarchical Pitman-Yor processes (HPYPs) and neural networks (NNs).

%For now, keep with tags and word separately, but note exceptions...
Our model defines a joint probability distribution 
%$p(\mathbf{t}_{1:n}, \mathbf{w}_{1:n}, \mathbf{a}_{1:2n})$ 
over a parsed sentence with POS tags $\mathbf{t}_{1:n}$, words $\mathbf{w}_{1:n}$
and a transition sequence $\mathbf{a}_{1:2n}$ as 
\begin{align*}
& p(\mathbf{t}_{1:n}, \mathbf{w}_{1:n}, \mathbf{a}_{1:2n}) \\
&= \prod_{i=1}^{n} \Big( p(t_i| \mathbf{h}^t_{m_i}) p(w_i|t_i, \mathbf{h}^w_{m_i}) \prod_{j = m_i + 1}^{m_{i+1}}
p(a_j | \mathbf{h}^a_{j}) \Big),
\end{align*}  %include marginalising derivations?
where $m_i$ is the number of transitions that have been performed when
$(t_i, w_i)$ is generated and $\mathbf{h}^t, \mathbf{h}^w$ and $\mathbf{h}^a$ 
are sequences representing the conditioning contexts for the tag, word and transition 
distributions, respectively. 
%Fixed-length templates for the conditioning contexts,
%based on elements in the current parser configuration at each prediction,
%are defined for each of our models.

In the generative process a shift transition is followed by a sequence of $0$ or
more reduce transitions. 
This is repeated until all the words
have been generated and a terminal configuration of the parser has been
reached. 
We shall also consider unlexicalised models, based only on POS tags.

%If the root symbol is put at the end of the sentence, once it 
%has been generated the other words on the stack have be to added 
%as children of the root with left-arcs. 

%In addition to considering models that generate both words and POS tags,
%we also consider models where only words are generated, or unlexicalised 
%models with only tags.

%training objectives
%For supervised training, given a training set of parsed and tagged sentences,
%we use the oracle to extract a derivation (transition sequence) for each 
%sentence. From each derivations we extract sets of training examples
%(predictions and context tuples) for our tag prediction, word prediction and
%transition prediction models. 

%First check if this really helps...
%For language modelling, the training objective is modified such that we allow
%any transition sequence whose arcs are a subset of the arcs in the training 
%data. 

%still too long
\subsection{Hierarchical Pitman-Yor processes}

The probability distributions for predicting words, tags and transitions 
are drawn from hierarchical Pitmar-Yor Process (HPYP) priors.
HPYP models were originally proposed for $n$-gram language 
modelling~\cite{Teh06}, %also Goldwater
and have been applied to various NLP tasks. %cite
A version of approximate inference in the HPYP model recovers interpolated 
Kneser-Ney smoothing~\cite{KneserNey95}, one of the best preforming $n$-gram 
language models. 
%Do we want this here or rather later
The Pitman-Yor Process (PYP) is a generalization of the Dirichlet process which
defines a distribution over distributions over a probability space 
$X$, with discount parameter $0 \le d < 1$, strength parameter $\theta > -d$ and 
base distribution $B$. 
PYP priors encode the power-law distribution found in 
the distribution of words. 

Sampling from the posterior is 
characterized by the Chinese Restaurant Process analogy, where  
each variable in a sequence is represented by a customer entering a 
restaurant and sitting at one of an infinite number of tables. 
Let $c_k$ be the number of customers sitting at table $k$ and $K$ the number 
of occupied tables. 
The customer chooses to sit at a table according to the probability 
\begin{equation*}
P(z_i = k | \textbf{z}_{1:i-1}) = \left\{
  \begin{array}{ll}
  \frac{c_k - d}{i - 1 + \theta} & 1 \le k \le K \\
  \frac{Kd + \theta}{i - 1 + \theta} & k = K + 1,
  \end{array} \right.
\end{equation*}
where $z_i$ is the index of the table chosen by the $i$th customer and
$\textbf{z}_{1:i-1}$ is the seating arrangement of the previous
$i-1$ customers.

All customers at a table share the same dish, corresponding to the
value assigned to the variables they represent. When a customer 
sits at an empty table, a dish is assigned to the table by 
drawing from the base distribution of the PYP. 

For HPYPs, the PYP base distribution can itself be drawn from a PYP.
The restaurant analogy is extended to the Chinese Restaurant Franchise,
where the base distribution of a PYP corresponds to another restaurant.
So when a customer sits at a new table, the dish is chosen by letting
a new customer enter the base distribution restaurant. All dishes
can be traced back to a uniform base distribution at the top of the hierarchy.

Inference over seating arrangements in the model is performed with Gibbs
sampling, based on routines to add or remove a customer from a 
restaurant. 
In our implementation we use the efficient data structures proposed by 
\newcite{BlunsomCGJ09}.
In addition to sampling the seating arrangement, the discount and strength
parameters are also sampled, using slice sampling. 
%TODO say something about particle filtering iff you use it

%Draws from the generative process are defined as 
%\begin{align*}
%    & t | \mathbf{h}^t & \sim\ &  T_{\mathbf{h}^t} \\
%    & w | \mathbf{h}^w & \sim\ &  W_{\mathbf{h}^w} \\
%    & a | \mathbf{h}^a & \sim\ &  A_{\mathbf{h}^a} 
%\end{align*}

In our model $T_{\mathbf{h}^t}, W_{\mathbf{h}^w}$ and $A_{\mathbf{h}^a}$ 
are HPYPs for the tag, word and transition distributions, respectively.  
The PYPs for the transition prediction distribution, with conditioning
context sequence $\mathbf{h}^a_{1:L}$, are defined hierarchically as
\begin{align*}
    & A_{\mathbf{h}^a_{1:L}} & \sim\ & \textrm{PYP}(d^A_L, \theta^A_L, 
          A_{\mathbf{h}^a_{1:L-1}}) \\ 
    & A_{\mathbf{h}^a_{1:L-1}} & \sim\ & \textrm{PYP}(d^A_{L-1}, \theta^A_{L-1}, 
          A_{\mathbf{h}^a_{1:L-2}}) \\ 
    & \ldots & \ldots & \\
& A_{\emptyset} & \sim\ &  \textrm{PYP}(d^A_0, \theta^A_0, \textrm{Uniform}),  
\end{align*}
where $d^A_k$ and $\theta^A_k$ are the discount and strength
discount parameters for PYPs with conditioning context 
length $k$. 
Each back-off level drops one context element.
The distribution given the empty context backs off to the uniform distribution
over all predictions. 
The word and tag distributions are defined by 
similarly-structured HPYPs.

The prior specifies an ordering of the symbols in the
context from most informative to least informative to the distributions being
estimated. 
The choice and ordering of this context is crucial in the 
formulation of our model. %, in order to avoid sparsity in the back-off structure.
%As words are more sparse than POS tags, they are always dropped first in the
%back-off structure. 
The contexts that we use %and arc-eager
are given in Table \ref{tab-pyp-ctx}.
%%this needs to be backed up experimentally, now!

%For unlexicalised parsing, the word contexts are excluded. 

%have different sets of contexts?
\begin{table}[t]
\centering
\begin{tabular}{|l|l|}
\hline
& Context elements \\
\hline
$ a_i $ & $ \sigma_1.t, \sigma_2.t, rc_1(\sigma_1).t, lc_1(\sigma_1).t, \sigma_3.t, $ \\
        & $ rc_1(\sigma_2).t, \sigma_1.w, \sigma_2.w $ \\
\hline
$ t_j $ & $ \sigma_1.t, \sigma_2.t, rc_1(\sigma_1).t, lc_1(\sigma_1).t, \sigma_3.t, $ \\
        & $ rc_1(\sigma_2).t, \sigma_1.w, \sigma_2.w $ \\
\hline
$ w_j $ & $ \beta.t, \sigma_1.t, rc_1(\sigma_1).t, lc_1(\sigma_1).t, \sigma_1.w, \sigma_2.w $ \\  %s1.t not really included
\hline
\end{tabular}

\caption{HPYP prediction contexts for the transition, tag and word distributions.
The context elements are ordered from most important to least important; 
the last elements in the lists are dropped first in the back-off structure.
The POS tag of node $s$ is referred to as $s.t$ and the word type as $s.w$.}
\label{tab-pyp-ctx}
\end{table}

%% file: inference.tex
\section{Decoding} %Inference
\label{sec-inference}
%\subsection{Decoding}

%TODO update for direction not-deterministic, predict tags
\begin{algorithm}[t]
 \footnotesize  %else \small
 \KwIn{Sentence $\mathbf{w}_{1:n}$, $K$ particles.}
 \KwOut{Parse tree of $ \argmax_{d \textrm{ in beam}} d.\theta$.}
 \emph{Initialize} the beam with parser configuration $d$ with weight $d.\theta = 1$\
 and $d.k = K$ particles\;
 \For{$i \leftarrow 1$ \KwTo $N$} {
     \emph{Search step}\;
     \ForEach{derivation $d$ in beam} {  
         $\textrm{nShift} = \textrm{round}(d.k \cdot p(\textrm{sh} | d.\mathbf{h}^{a}))$\;
         $\textrm{nReduce} = d.k - \textrm{nShift}$\;
         \If{$\textrm{nReduce} > 0$} {
             $a = \argmax_{a \ne \textrm{sh}} p(a | d.\mathbf{h}^{a})$\;
             beam.append($dd \leftarrow d$)\;
             $dd.k \leftarrow \textrm{nReduce}$\;
             $dd.\theta \leftarrow dd.\theta \cdot p(a | d.\mathbf{h}^{a})$\;
             $dd$.execute($a$)\;
         } 
         $d.k \leftarrow \textrm{nShift}$\;
         \If{$\textrm{nShift} > 0$} {
             $d.\theta \leftarrow d.\theta \cdot p(\textrm{sh} | d.\mathbf{h}^{a}) \cdot \max_{t_i} 
                                           p(t_i | d.\mathbf{h}^{t}) 
                                           p(w_i | d.\mathbf{h}^{w})$\;
             $d$.execute(sh)\;
         }
     }
     \emph{Selection step}\;
     \ForEach{derivation $d$ in beam} {  
         $d.\theta' \leftarrow \frac {d.k \cdot d.\theta} {\sum_{d'} d'.k \cdot d'.\theta}$\;
     }
     \ForEach{derivation $d$ in beam} {  
         $d.k = \lfloor d.\theta' \cdot K \rfloor $\;
         \If{$d.k = 0$} {
             beam.remove($d$)\;    
         }
     }
 }
 %\emph{bestDerivation} $ \leftarrow \argmax_{d \textrm{ in beam}} d.\theta$\;
\caption{Beam search decoder for arc-standard generative dependency parsing.}
 \label{alg:beam-search}
\end{algorithm}

In the standard approach to beam search for transition-based 
parsing~\cite{ZhangC08}, the beam stores partial derivations with the same 
number of transitions performed, and the lowest-scoring ones are 
removed when the size of the beam exceeds a set threshold. 
However, in our model we cannot compare derivations with the same number of
transitions but which differ in the number of words shifted. 
One solution is to keep $n$ separate beams, each containing only derivations 
with $i$ words shifted, but this approach leads to $O(n^2)$ decoding complexity. 
Another option is to prune the beam every time after the next word is shifted in
all derivations -- however the number of reduce transitions
that can be performed between shifts is bounded by the stack size, so decoding 
complexity remains quadratic.

We propose a novel linear-time decoding algorithm inspired by particle
filtering (see Algorithm \ref{alg:beam-search}).
Instead of specifying a fixed limit on the size of the beam, the beam size is
controlled by setting the number of particles $K$. Every partial derivation
$d_j$ in the beam is associated with $k_j$ particles, such that $\sum_j k_j = K$.
Each pass through the beam advances each $d_j$ until the next word is
shifted.
%The main idea of the beam search is that after the $i$th pass through the
%beam, $i$ words have been generated in each of the derivations on the beam.

At each step, to predict the next transition for $d_j$, $k_j$ is divided 
proportionally between taking a shift or reduce transition, according to 
$p(a | d_j.\mathbf{h}^a)$. 
If a non-zero number of particles are assigned to reduce, the highest 
scoring left-arc and right-arc transitions are chosen deterministically, 
and derivations that execute them are added to the beam. In practice
we found that adding only the highest scoring reduce transition
(left-arc or right-arc) gives very similar performance. 
The shift transition is performed on the current derivation, and the
derivation weight is also updated with the word generation probability. 

A POS tag is also generated along with a shift transition. Up to three
candidate tags are assigned (more do not improve performance)
and corresponding derivations are added to the
beam, with particles distributed relative to the tag probability (in Algorithm
\ref{alg:beam-search} only one tag is predicted).

A pass is complete once the derivations in the beam, including those added 
by reduce transitions during the pass, have been iterated through.
Then a selection step is 
performed to determine which derivations are kept.
The number of particles for each derivation are reallocated based on the 
normalised weights of the derivations, each weighted by its current number of 
particles. Derivations to which zero particles are assigned are eliminated.
The selection step allows the size of the beam to depend on the uncertainty of
the model during decoding. 
The selectional branching method proposed by 
\newcite{ChoiM13} for discriminative beam-search parsing has a similar goal.

After the last word in the sentence has been shifted, reduce transitions
are performed on each derivation until it reaches a terminal configuration. 
The parse tree corresponding to the highest scoring final derivation is returned. 
%TODO update if root-final has been cleared up
%mention enforcement of 1 root word (if necessary for parsing)? 

The main differences between our algorithm and particle filtering are
that we divide particles proportionally instead of sampling with replacement,
and in the selection step we base the redistribution on the derivation weight instead of 
the importance weight (the word generation probability). Our method can be
interpreted as maximizing by sampling from a peaked version of the distribution over 
derivations.

%% file: experiments.tex
%%remove question parsing, add langauge modelling
\section{Experiments}
\label{sec-experiments}

\subsection{Parsing Setup}

We evaluate our model as a parser on the standard
English Penn Treebank~\cite{MarcusSM93} setup, training
on WSJ sections 02-21, developing on section 22, and testing 
on section 23. 
We use the head-finding rules of 
\newcite{YamadaM03} (YM)\footnote{http://stp.lingfil.uu.se/~nivre/research/Penn2Malt.html}
for constituency-to-dependency conversion, to enable comparison with previous
results. 
We also evaluate on the Stanford dependency 
representation~\cite{DeMarneffeM08}
(SD)\footnote{Converted with version 3.4.1 of the Stanford parser,
available at http:/nlp.stanford.edu/software/lex-parser.shtml.}.
%We use the option to treat copular verbs as sentence heads,
%which produces more syntactically-motivated dependencies.
%the projective CoNLL syntactic dependency 
%representation\footnote{Converted with the LTH converter,
%ignoring function tags and null elements.
%Available at http://nlp.cs.lth.se/software/treebank\_converter/}
%(CD)~\cite{JohanssonN07}, and 

%Question Parsing

%For Stanford dependencies we consider the default version, as well as
%the version in which copular verbs are treated as the head of the sentence,
%which This produces more syntactically-motivated dependency 
%structures\footnote{The other main difference between CD and SD parses lies in
%the choice of head word between an auxiliary and main lexical verb in a clause.}. 

Words that occur only once in the training data are treated as unknown words. 
We classify unknown words according to capitalization, numbers, punctuation 
and common suffixes into classes similar to those used in the implementation 
of generative constituency parsers such as the Stanford parser~\cite{KleinM03}.

As a discriminative baseline we use MaltParser~\cite{NivreHN06}, a discriminative, greedy
transition-based parser, performing arc-standard parsing with LibLinear as classifier.
Although the accuracy of this model is not state-of-the-art, it does
enable us to compare our model against an optimised discriminative model with a 
feature-set based on the same elements as we include in our conditioning contexts.

Our HPYP dependency parser (HPYP-DP) is trained with $20$ iterations of Gibbs sampling, 
resampling the hyper-parameters after every iteration, except when
performing inference over latent structure, in which case they are only resampled 
every $5$ iterations. 
Training with a deterministic oracle takes
$28$ seconds per iteration (excluding resampling hyper-parameters),
while a non-deterministic oracle (sampling with 
$100$ particles) takes $458$ seconds.

\subsection{Modelling Choices}

\begin{table}
\centering
\begin{tabular}{|l|c|c|}
\hline
Model & UAS & LAS \\
\hline 
MaltParser Unlex & 85.23 & 82.80 \\ 
%MaltParser 10k & 88.14 & 86.94 \\
MaltParser Lex & 89.17 & 87.81 \\ 
%\newcite{ChenM14} & 89.09 & 87.72 \\ %still not convinced
\hline
Unlexicalised & 85.64 & 82.93  \\ 
%Lexicalised 10k & 87.27 & 84.68 \\
Lexicalised, unlex context & 87.95 & 85.04 \\ 
Lexicalised, tagger POS & 87.84 & 85.54 \\
\textbf{Lexicalised, predict POS} & \textbf{89.09} & \textbf{86.78} \\
Lexicalised, gold POS & 89.30 & 87.28 \\ 
%HPYP discriminative (greedy) & 89.12 & 87.92  \\ %TODO
%HPYP discriminative (beam-search) & 90.08 & 88.76 \\
%sampling instead of max accuracy
\hline
\end{tabular}
\caption{HPYP parsing accuracies on the YM development set, for various
lexicalised and unlexicalised setups.}
\label{tab-ym-dev}
\end{table}

We consider several modelling choices in the construction
of our generative dependency parsing model. Development set parsing results 
are given in Table \ref{tab-ym-dev}. 
We report unlabelled attachment score (UAS) and labelled attachment
score (LAS), excluding punctuation.  
%TODO MaltParser as a baseline, even though a weak one...
%Also consider zpar if we can modify feature sets

\subsubsection*{HPYP priors}

The first modelling choice is the selection and ordering of 
elements in the conditioning contexts of the HPYP priors. 
Table \ref{tab-context} shows how the development set accuracy
increases as more elements are added to the conditioning context. 
The first two words on the stack are the most important, but insufficient -- 
second-order dependencies and further elements on the stack should also
be included in the contexts. 
The challenge is that the back-off structure of each HPYP specifies an
ordering of the elements based on their importance in the prediction.
We are therefore much more restricted than classifiers with large, sparse
feature-sets which are commonly used in transition-based parsers.
Due to sparsity, the word types are the first elements to be dropped in the
back-off structure, and elements such as third-order dependencies, which have been shown 
to improve parsing performance, cannot be included successfully in our model. 

\begin{table}
\centering
\begin{tabular}{|l|c|c|}
\hline
    Context elements         &  UAS  &  LAS  \\           
    \hline    
    $\sigma_1.t, \sigma_2.t$ & 73.25 & 70.14 \\
    $+rc_1(\sigma_1).t$      & 80.21 & 76.64 \\
    $+lc_1(\sigma_1).t$      & 85.18 & 82.03 \\
    $+\sigma_3.t$            & 87.23 & 84.26 \\ 
    $+rc_1(\sigma_2).t$      & 87.95 & 85.04 \\
    $+\sigma_1.w$            & 88.53 & 86.11 \\  
    $+\sigma_2.w$            & 88.93 & 86.57 \\
    \hline
\end{tabular}
\caption{Effect of including elements in the model conditioning contexts.
Results are given on the YM development set.}
\label{tab-context}
\end{table}

Sampling over parsing derivations during training further improves performance
by $0.16\%$ to $89.09$ UAS. 
%Also report root accuracy
Adding the root symbol at the end of the sentence rather than at the front 
gives very similar parsing performance. %, but does not seem to be able to model
%sentence length as well. 
When unknown words are not clustered according to surface
features, performance drops to $88.60$ UAS. 

%Restricting the parser to predict only one head word 
%per sentence improves performance by $0.2$ UAS %check
%against allowing the model to predict multiple heads.

\subsubsection*{POS tags and lexicalisation}

It is standard practice in transition-based parsing to obtain POS tags with
a stand-alone tagger before parsing. 
However, as we have a generative model, we can use the model to assign POS tags 
in decoding, while predicting the transition sequence. We compare 
predicting tags against using gold standard POS tags and tags obtain using the 
Stanford POS tagger\footnote{
We use the efficient ``left 3 words'' model, 
trained on the same data as the parsing model, excluding distributional features. 
Tagging accuracy is $95.9\%$ on the development set and $96.5\%$ on the test
set.}~\cite{ToutanovaKMS03}.
Even though the predicted tags are slightly less accurate than the Stanford
tags on the development set ($95.6\%$), jointly predicting tags and decoding
increases the UAS by $1.1\%$. 
The jointly predicted tags are a better fit to the generative model, which can be seen by
an improvement in the likelihood of the test data.
\newcite{BohnetN12} found that joint prediction increases both POS and 
parsing accuracy. However, their model rescored a $k$-best list of tags 
obtained with an preprocessing tagger, while our model does not use the 
external tagger at all during joint prediction. 

We train lexicalised and unlexicalised versions of our model.
Unlexicalised parsing gives us a strong baseline ($85.6$ UAS)
over which to consider our model's ability to predict and condition on words. 
Unlexicalised parsing is also considered to be robust for applications 
such as cross-lingual parsing~\cite{McdonaldPH11}.
Additionally, we consider a version of the model that don't include
lexical elements in the conditioning context. 
%The probabilistic model then factors as 
%$p(\mathbf{t}_{1:n}, \mathbf{a}_{1:2n}) \prod_{i=1}^n p(w_i|\mathbf{t}_{1:i}, 
%\mathbf{a}_{1:m_i})$.
This model performs only $1\%$ UAS lower than the best lexicalised model, although
it makes much stronger independence assumptions. The main benefit of lexicalised
conditioning contexts are to make incremental decoding easier. 

%Motivate
%not really useful...
%\subsubsection*{Sentence roots}

\subsubsection*{Speed vs accuracy trade-offs}

\begin{table}
\centering
\begin{tabular}{|c|c|c|}
\hline
Particles & Sent/sec & UAS \\
\hline
%Beam 32 & 3 & 88.82 \\
%Beam 16 & 7 & 88.46 \\
5000  & 18 & 89.04 \\
1000  & 27 & 88.93 \\
100  & 54 & 87.99 \\
10  & 104 & 85.27 \\
\hline
%Beam 16 & 29 & 87.64 \\
1000 & 108 & 87.59 \\
100  & 198 & 87.46 \\
10   & 333 & 85.86 \\    
\hline
\end{tabular}
\caption{Speed and accuracy for different configurations of the 
    decoding algorithm. Above the line, POS tags are predicted by the model, 
below pre-tagged POS are used.}
    %Entries with ``beam'' use standard beam-search,
    %the rest use our particle filter-based decoding algorithm.}
    \label{tab-speed-acc}
\end{table}

We consider a number of trade-offs between speed and accuracy in the model. 
We compare using different numbers of particles during decoding, as well as
jointly predicting POS tags against using pre-obtained tags (Table \ref{tab-speed-acc}). 

The optimal number of particles is found
to be $1000$ - more particles only increase accuracy by about $0.1$ UAS. 
Although jointly predicting tags is more accurate, using pre-obtained tags
provides a better trade-off between speed and accuracy -- $87.59$ against
$85.27$ UAS at around $100$ sentences per second. In comparison, the 
MaltParser parses around $500$ sentences per second. 

We also compare our particle filter-based algorithm against a more standard
beam-search algorithm that prunes the beam to a fixed size after each word is
shifted. This algorithm is much slower than the particle-based algorithm
-- to get similar accuracy it parses only $3$ sentences per second
(against $27$) when predicting tags jointly, and $29$ (against $108$)
when using pre-obtained tags.

%his shows that the distribution over arcs is strongly peaked, and 
%most of the ambiguity in the model during decoding lies in the 
%choice between shifting and reducing. 

%estimate search error?

\subsection{Parsing Results}

\begin{table}
\centering
\begin{tabular}{|l|c|c|}
\hline  
Model & UAS & LAS \\
\hline %generative 
\newcite{Eisner96} & 80.7 & - \\
\newcite{WallachSM08} & 85.7 & - \\
\newcite{TitovH07} & 89.36 & 87.65 \\  
\hline  
\textbf{HPYP-DP} & \textbf{88.47} & \textbf{86.13} \\
\hline %greedy
MaltParser & 88.88 & 87.41 \\  
%\newcite{ChenM14}\dag & 88.84 & 87.29 \\ %still not convinced
  %90.20 & 89.10 \\  % Daume ea  
%\newcite{GoldbergN13} & 91.84 & 90.20 \\ 
%\hline %beam-search
%\newcite{ZhangN11} \dag & 91.79 & 90.58 \\
%\newcite{ZhangC08} & 92.1 & - \\
%\newcite{HuangS10} & 92.1 & \\
\newcite{ZhangN11} & 92.9 & 91.8 \\
%\newcite{BohnetN12} & 93.38 & 92.44 \\
\newcite{ChoiM13} & 92.96 & 91.93 \\
%\newcite{MartinsAS13}\dag (2nd order) & 92.06 & 90.81 \\ 
%\newcite{MartinsAS13} & 93.07 & - \\
\hline
\end{tabular}
\caption{Parsing accuracies on the YM test set.
compared against previous published results.
\newcite{TitovH07} was retrained to
enable direct comparison.}
\label{tab-ym-test}
\end{table}

Test set results comparing our model against existing discriminative and generative 
dependency parsers are given in Table \ref{tab-ym-test}.
Our HPYP model performs much better than Eisner's 
generative model as well as the Bayesian version of that model proposed
by \newcite{WallachSM08} (the result for Eisner's model is given as
reported by \newcite{WallachSM08} on the WSJ). 
The accuracy of our model is only $0.8$ UAS below the generative model of
\newcite{TitovH07}, despite that model being much more powerful.
The Titov and Henderson model takes $3$ days to train, and its decoding
speed is around $1$ sentence per second.

The UAS of our model is very close to that of the MaltParser. 
%and the neural network parser of \newcite{ChenM14}\footnote{Trained without 
%initializing with words embeddings obtained by pre-training on a much larger external dataset.}.
However, we do note that our model's performance is relatively worse on LAS
than on UAS. An explanation for this is that as we do not include labels in the 
conditioning contexts, the predicted labels are independent of words that have not yet 
been generated.

We also test the model on the Stanford dependencies, which have a larger
label set. 
Our model obtains 
$87.9$/$83.2$ against the MaltParser's $88.9$/$86.2$ UAS/LAS. 

Despite these promising results, our model's performance still lags behind 
recent discriminative parsers~\cite{ZhangN11,ChoiM13} with beam-search and richer
feature sets than can be incorporated in our model. 
In terms of speed,
\newcite{ZhangN11} parse $29$ sentences per second, against the $110$ sentences per second
of \newcite{ChoiM13}. 
Recently proposed neural networks for dependency parsers have further 
improved performance~\cite{DyerEa15,WeissACP15}, reaching up to $94.0\%$ UAS
with Stanford dependencies. 

We argue that the main weakness of the
HPYP parser is sparsity in the large conditioning contexts composed of tags and words. 
The POS tags in the parser configuration context
already give a very strong signal for predicting the next transition.
As a result it is challenging to construct PYP reduction lists
that also include word types without making the back-off contexts too sparse. 

The other limitation is that our decoding algorithm, although efficient, still
prunes the search space aggressively, while not being able to take advantage of 
look-ahead features as discriminative models can. 
Interestingly, we note that a discriminative parser cannot reach high
performance without look-ahead features. %TODO
% even when a large beam is used. %evidence

%TODO decoding speeds (test only?)
\subsection{Language Modelling}

%with a unlabelled version of our parser.
%The end-of-sentence symbol is predicted implicitly when
%only the root symbol is left on the stack. 

Next we evaluate our model as a language model.
First we use the standard WSJ language modelling 
setup, training on sections $00-20$, developing on
$21-22$ and testing on $23-24$. 
Punctuation is removed, numbers and symbols are 
mapped to a single symbol and the vocabulary is limited
to $10,000$ words. 
Second we consider a semi-supervised setup where we train the model, in addition to the
WSJ, on a subset of $1$ million sentences ($24.1$ million words) from the WMT English
monolingual training data\footnote{Available at http://www.statmt.org/wmt14/translation-task.html.}.
%Note that this is only a portion of the total training data. 
This model is evaluated on \texttt{newstest2012}.

When training our models for language modelling, we first perform standard
supervised training, as for parsing (although we don't predict labels).
This is followed by a second training stage, where we train the model only on words, 
regarding the tags and parse trees as latent structure. 
In this unsupervised stage we train the model with %$3$ iterations of 
particle Gibbs sampling~\cite{AndrieuDH10},
using a particle filter to sample parse trees.
When only training on the WSJ, we perform this step on the same data, now
allowing the model to learn parses that are not necessarily consistent with the
annotated parse trees. 

For semi-supervised training, unsupervised learning is performed on the large
unannotated corpus.
However, here we find the highest scoring parse trees, rather than sampling. 
Only the word prediction distribution is updated, not the tag and 
transition distributions. 

Language modelling perplexity results are given in Table \ref{tab-wsj-lm}.
We note that the perplexities reported are
upper bounds on the true perplexity of the model,
as it is intractable to sum over all 
possible parses of a sentence to compute the
marginal probability of the words. As an 
approximation we sum over the final beam 
after decoding.

The results show that on the WSJ the model performs slightly better than
a HPYP $n$-gram model. One disadvantage of evaluating on this dataset is that 
due to removing punctuation and restricting the vocabulary, 
the model parsing accuracy drops to $84.6$ UAS. Also note that in contrast to many 
other evaluations, we do not interpolate with a $n$-gram model -- this will improve
perplexity further. 

On the big dataset we see a larger improvement over the $n$-gram
model. This is a promising result, as it shows that our model can successfully
generalize to larger vocabularies and unannotated datasets. 
%For future work we plan to train the model on even larger datasets.

\begin{table}  
\centering
\begin{tabular}{|l|c|}
\hline
Model & Perplexity \\
\hline
%SRILM $5$-gram & 147.86 \\ %reported 141.2
HPYP $5$-gram & 147.22 \\ 
\newcite{ChelbaJ00} & 146.1 \\
\newcite{EmamiJ05}  & 131.3 \\
\textbf{HPYP-DP} & \textbf{145.54} \\ %can we do better?
\hline
HPYP $5$-gram & 178.13 \\
\textbf{HPYP-DP} & \textbf{163.96} \\
\hline
\end{tabular}
\caption{Language modelling test results. Above, training and testing on WSJ.
Below, training semi-supervised and testing on WMT.}
\label{tab-wsj-lm}
\end{table}

\subsection{Generation}

To support our claim that our generative model is a good model for 
sentences, we generate some examples. 
The samples given here were obtained by generating $1000$ samples, and choosing
the $10$ highest scoring ones with length greater or equal to $10$. The models are trained on
the standard WSJ training set (including punctuation). 

The examples are given in Table \ref{tab-gen}. 
The quality of the sentences generated by the dependency model is superior to that 
of the $n$-gram model, despite the models have similar test set 
perplexities.
The sentences generated by the dependency model tend to have more global syntactic structure
(for examples having verbs where expected), while retaining the local coherence of $n$-gram models.
The dependency model was also able to generate balanced quotation marks. 

\begin{table*}
\small    
\centering
\begin{tabular}{|l|}
\hline    
sales rose NUM to NUM million from \$ NUM . \\
estimated volume was about \$ NUM a share , . \\
meanwhile , annual sales rose to NUM \% from \$ NUM . \\
mr. bush 's profit climbed NUM \% , to \$ NUM from \$ NUM million million , or NUM cents a share . \\
treasury securities inc. is a unit of great issues . \\
`` he is looking out their shareholders , '' says . \\
while he has done well , she was out . \\
that 's increased in the second quarter 's new conventional wisdom . \\
mci communications said net dropped NUM \% for an investor . \\
association motorola inc. , offering of \$ NUM and NUM cents a share . \\ 
\hline
otherwise , actual profit is compared with the 300-day estimate . \\
the companies are followed by at least three analysts , and had a minimum five-cent change in actual earnings per share . \\
bonds : shearson lehman hutton treasury index NUM , up  \\
posted yields on NUM year mortgage commitments for delivery within NUM days .  \\
in composite trading on the new york mercantile exchange .  \\
the company , which has NUM million shares outstanding .  \\
the NUM results included a one-time gain of \$ NUM million . \\ 
however , operating profit fell NUM \% to \$ NUM billion from \$ NUM billion .  \\
merrill lynch ready assets trust : NUM \% NUM days ; NUM \% NUM to NUM days ; NUM \% NUM to NUM days . \\
in new york stock exchange composite trading , one trader .  \\
\hline
\end{tabular}    
\caption{Sentences generated, above by the generative dependency model, below by a $n$-gram model. 
In both cases, $1000$ samples were generated, and the most likely sentences of length $10$ or more
are given.}
\label{tab-gen}
\end{table*}

\begin{comment}
\subsection{Multilingual parsing}

%Is CTB really necc.?
We evaluate our models on the CoNLL 2006 multilingual dependency
parsing datasets~\cite{BuchholzM06}, as well as the CoNLL 2008 non-projective
English syntactic dependencies, %cite
for both parsing and language modelling.
We report the parsing results in table \ref{tab-multilingual}.

%We also evaluate on the Chinese Treebank (CTB).
%We compare against the results of %XX.

%TODO Results

\begin{table}
\centering
\begin{tabular}{|l|c|c|}
\hline
Language & NN (UAS) & TurboParser (UAS) \\
\hline
Arabic      & 77.0 & 79.6 \\
%Basque      & 73.7 & 
Bulgarian   & 88.6 & 93.1 \\
%Catalan     & 88.2 & 
Chinese     & 85.1 & 89.9 \\
Czech       & 77.6 & 90.3 \\
Danish      & 85.8 & 91.5 \\
Dutch       & 85.4 & 86.2 \\
English     & 87.8 & 93.2 \\
German      & 84.4 & 92.4 \\
%Greek       & 79.8 & 
%Hungarian   & 75.8 & 
%Italian     & 79.0 & 
Japanese    & 92.8 & 93.5 \\
Portuguese  & 88.4 & 92.7 \\
Slovene     & 76.7 & 86.0 \\
Spanish     & 75.6 & 85.6 \\
Swedish     & 85.1 & 91.4 \\
Turkish     & 75.5 & 76.9 \\
\hline
\end{tabular}
\caption{CoNLL multilingual parsing results.}
\label{tab-multilingual}

\end{table} 
\end{comment}

%% file: related-work.tex
\section{Related work}
\label{sec-related-work}

One of the earliest graph-based dependency parsing 
models~\cite{Eisner96} is generative, estimating the probability of 
dependents given their head and previously generated siblings. 
To counter sparsity in the conditioning context of the distributions, 
backoff and smoothing are performed. 
\newcite{WallachSM08} proposed a Bayesian HPYP parameterisation of this model.
%However, their unlabelled attachment score of $85.7\%$ on WSJ is lower than
%that our our HPYP model. %In experimental results? or include my result here

Other generative models for dependency trees have been proposed mostly 
in the context of unsupervised parsing. 
The first successful model was the dependency model with valence 
(DMV)~\cite{KleinM04}. Several extensions have been proposed for this
model, for example using structural annaeling~\cite{SmithE06}, %SpitkovskyVAJ10}, 
Viterbi EM training~\cite{SpitkovskyAJM10} or richer 
contexts~\cite{BlunsomC10}.    %HeaddenJM09,
However, these models are not powerful enough 
for either accurate parsing or language modelling with rich contexts
(they are usually restricted to first-order dependencies and valency).

%%syntactic LMs
Although any generative parsing model can be applied to language modelling
by marginalising out the possible parses of a sentence, in practice the success
of such models has been limited. Lexicalised PCFGs applied to
language modelling~\cite{Roark01,Charniak01} show improvements over $n$-gram
models, but decoding is prohibitively expensive for practical integration in 
language generation applications. 

\newcite{ChelbaJ00} as well as \newcite{EmamiJ05} proposed
incremental syntactic language models with some similarities to our
model.
Those models predict binarized constituency trees with a transition-based
model, and are parameterized by deleted interpolation and neural 
networks, respectively. 
\newcite{RastrowDK12} applies a transition-based dependency
language model to speech recognition, using hierarchical 
interpolation and relative entropy pruning. 
However, the model perplexity only improves over an $n$-gram model 
when interpolated with one. 

\newcite{TitovH07} introduced a generative latent variable model for 
transition-based parsing.
The model is based on an incremental sigmoid belief networks, using the
arc-eager parsing strategy.
Exact inference is intractable, so neural networks %feed-forward
and variational mean field methods are proposed to perform approximate inference.
However, this is much slower and therefore less scalable than our model.
%Other Henderson models

A generative transition-based parsing model for non-projective parsing
is proposed in \cite{CohenGS11}, along with a dynamic program for 
inference. The parser is similar to ours, but the dynamic program restricts 
the conditioning context to the top $2$ or $3$ words on the stack.
No experimental results are included. 

\newcite{LeZ14} proposed a recursive neural network generative model over
dependency trees. However, their model can only score trees, not perform parsing,
and its perplexity ($236.58$ on the PTB development set) is worse than model's,
despite using neural networks to combat sparsity.

%There has recently been interest in applying neural networks and other
%models based on distributed representations to dependency 
%parsing~\cite{ChenM14,LeiBJ14} as well as to learn distributed word 
%representations tailored to dependency 
%structures~\cite{LevyG14,BansalGL14}. %compare Chen in experiments

Finally, incremental parsing with particle filtering has been proposed 
previously~\cite{LevyRG09} to model human online sentence processing.